\documentclass[letterpaper]{article} 
\usepackage{aaai25}  
\usepackage{times}  
\usepackage{helvet}  
\usepackage{courier}  
\usepackage[hyphens]{url}  
\usepackage{graphicx} 
\usepackage{natbib}  
\usepackage{caption} 
\usepackage{array}

\usepackage{makecell}
\usepackage{amsmath,amssymb,amsfonts}
\usepackage{amsthm}
\usepackage{multirow}
\usepackage{caption}
\usepackage{subfigure}
\usepackage{pgfplots}

\usepackage[framemethod=tikz]{mdframed}
\definecolor{mygray}{RGB}{243,243,244}
\newmdenv[
  innertopmargin=0pt,
  backgroundcolor=mygray,
  linecolor=none,
  innerleftmargin=0pt,
  innerrightmargin=0pt,
  leftmargin=0pt
  ]{mymath}

\usepackage[ruled, linesnumbered,vlined]{algorithm2e}

\title{\texttt{Pioneer}: Physics-informed Riemannian Graph ODE for\\Entropy-increasing Dynamics}
\author{
    Li Sun\textsuperscript{\rm 1}\thanks{Corresponding Author: Li Sun}, 
    Ziheng Zhang\textsuperscript{\rm 1},
    Zixi Wang\textsuperscript{\rm 1},
    Yujie Wang\textsuperscript{\rm 1},
    Qiqi Wan\textsuperscript{\rm 1},
    Hao Li\textsuperscript{\rm 1},
    Hao Peng\textsuperscript{\rm 2},
    Philip S. Yu\textsuperscript{\rm 3}
}
\affiliations{
    \textsuperscript{\rm 1}North China Electric Power University, Beijing 102206, China\\
    \textsuperscript{\rm 2}Beihang University, Beijing 100191, China\\
    \textsuperscript{\rm 3}Department of Computer Science, University of Illinois at Chicago, IL 60607, USA\\
    ccesunli@ncepu.edu.cn; zhangzh@ncepu.edu.cn; psyu@uic.edu

}

\begin{document}

\maketitle

\begin{abstract}
Dynamic interacting system modeling is important for understanding and simulating real world systems.
The system is typically described as a graph, where multiple objects dynamically interact with each other  and evolve over time.
In recent years, graph Ordinary Differential Equations (ODE) receive increasing research attentions.
While achieving encouraging results, existing solutions prioritize the traditional Euclidean space, and neglect the intrinsic geometry of the system and physics laws, e.g., the principle of entropy increasing.
The limitations above motivate us to rethink the system dynamics from a fresh perspective of Riemannian geometry,
and pose a more realistic problem of physics-informed dynamic system modeling, considering the underlying geometry and physics law for the first time.
In this paper,  we present a novel physics-informed Riemannian graph ODE for a wide range of entropy-increasing dynamic systems (termed as Pioneer).
In particular, we formulate a differential system on the Riemannian manifold, where a manifold-valued graph ODE is governed by the proposed constrained Ricci flow, and a manifold preserving Gyro-transform aware of system geometry.
Theoretically, we report the provable entropy non-decreasing of our formulation, obeying the physics laws.
Empirical results show the superiority of  Pioneer on real datasets.
\end{abstract}

%
\begin{links}
    \link{Code}{https://github.com/nakks2/Pioneer}
\end{links}


\section{Introduction}
The dynamic interacting systems are ubiquitous in the real world, 
ranging from meteorology
and the spread of COVID \cite{www24CausalGraphODE}
to social networks \cite{icml23HOPE}.
Dynamic system modeling is important for understanding and simulating the real-world systems.
In the literature, 
Ordinary Differential Equations (ODEs) \cite{nips18NODE} show success in modeling continuous evolvement.
Given that these interacting systems are naturally represented as graphs, Graph Neural Networks (GNNs) \cite{nips18gat} are frequently employed alongside other methods to incorporate the intercorrelations.
Recently, graph ODEs are introduced and achieve encouraging results \cite{kdd21CoupledGraphODE}.
However, there are several important issues largely remaining open.

\begin{figure} 
\centering 
\subfigure[Euclidean Modeling]{
\includegraphics[width=0.475\linewidth]{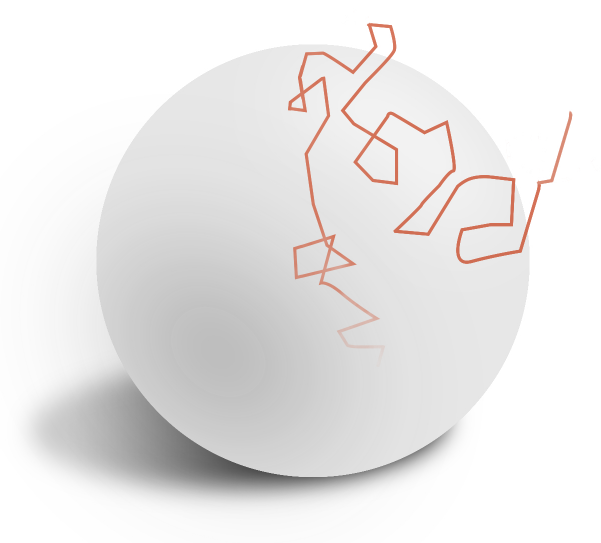}}
\subfigure[Riemannian Modeling]{
\includegraphics[width=0.485\linewidth]{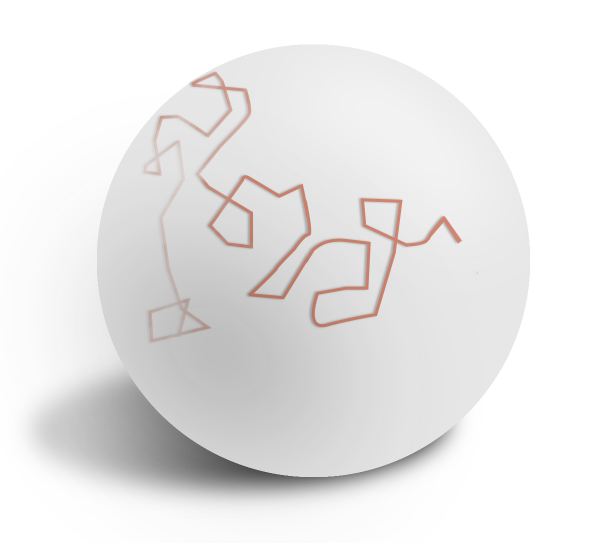}}
 \vspace{-0.05in}
\caption{Illustrated example on Earth's system. Traditional Euclidean models expose to generate unrealistic trajectory as in (a), while we preserve the manifold geometry as in (b).}
\label{Fig-example}
\end{figure}

On the one hand, previous studies prioritize the traditional Euclidean space, and neglect the intrinsic \textbf{geometry} of real systems.
We illustrate the importance of geometry awareness in Figure \ref{Fig-example}. 
Unaware of the geometry,  traditional Euclidean models generate trajectories that extend beyond the Earth's surface, which is unreasonable and unrealistic.
Riemannian geometry provides an elegant framework for complex systems \cite{aaai22sunli}. The Earth is a 2D sphere and does not have any isometric alternative in Euclidean space of any dimension \cite{o2009}.
Unfortunately, Riemannian manifold has not yet been introduced to model dynamic interacting system, to the best of our knowledge.

On the other hand, real systems obey the \textbf{physics laws}, e.g., the entropy-increasing principle of thermodynamics \cite{mackey1989}.
In contrast, existing solutions largely overlook such physics laws in dynamic modeling, 
and thus the generated states are not aligned with the physics law on real-world datasets, as investigated in our Experiment.
In recent years, physics-informed neural nets \cite{liu24,nips22PNNPDE} are to embed physics laws to the neural model, 
but they primarily focus on solving Partial Differential Equations (PDEs), 
and cannot use for regression or dynamical prediction yet.
In Riemannian geometry, Ricci flow is off the shelf to describe system evolvement \cite{Bai20,Ollivier2007}, 
but it is not aligned with the entropy-increasing principle  \cite{baptista2024}.
So far, it still lacks a principled way to incorporate the physics law with the dynamic models.

That is, \textbf{existing solutions expose to the danger of generating geometry-unaware or anti-physics trajectories.}
Such limitations result in unreasonable prediction contrary to the fact, preventing the wide use in real scenarios, 
and motivate us to rethink the dynamics modeling of real-world systems.
We argue that the success of dynamics modeling is characterized not only by prediction accuracy comparing to the truth, but also by the reasonable prediction in line with physics laws.
Thus, we pose a more realistic yet challenging problem of \emph{physics-informed dynamic system modeling}.

We approach the dynamic system modeling from a fresh perspective using Riemannian geometry,
and propose a novel \underline{P}hysics-\underline{i}nf\underline{o}rmed Riema\underline{n}nian graph OD\underline{E} for the ubiquitous  \underline{e}ntropy-inc\underline{r}easing dynamic systems, named \textbf{\texttt{Pioneer}}.
Given trajectory observations,
we model the underlying dynamics of the objects' co-evolvement, 
where the correlation among objects tends to evolve over time as well.
The key novelty is that, 
in light of aforementioned issues, 
\textbf{\texttt{Pioneer}} is also designed to generate reasonable outputs, ensuring that the predicted trajectories align with the system's geometry (such as the Earth's surface) and that the system states over time adhere to physical principles.
Specifically, \texttt{Pioneer} is a differential system on the Riemannian manifold,
where  a \textbf{Manifold ODE} is introduced to model the continuous co-evolvement with a  manifold-valued GNN.
Informed of physics laws,
a \textbf{constrained Ricci flow} is formulated  to study the continuous changes of object correlation.
In particular, 
we design a neural model over object features to constrain the canonical Ricci flow so that, 
\emph{theoretically, the proposed flow is proved to be entropy non-decreasing} (Theorem 3.1).
Being aware of the geometry, 
we propose \textbf{Gyro-transform} for encoding/decoding process, 
so that the  generated trajectories lie on the manifold.
Concretely, \emph{Gyro-transform conducts manifold-preserving transformation by simple matrix-vector multiplication} (Theorem 3.2),
and no longer uses the expensive exponential/logarithmic maps.
Finally, both accuracy and rationale are evaluated on the real datasets.

Overall, key contributions are summarized as follows:
\begin{itemize}
\item We rethink system dynamics over the graph, and pose \emph{a more realistic problem of physics-informed system dynamics},  considering underlying geometry and physics law for the first time, to the best of our knowledge.
\item We propose a physics-informed ordinary differential system on the Riemannian manifold (\textbf{\texttt{Pioneer}}), where we formulate a constrained Ricci flow, in line with physics law,  and manifold-preserving Gyro-transform. 
\item Extensive empirical results show \texttt{Pioneer} outperforms the state-of-the-art methods on several real datasets, and we analyze the geometry and entropy of real systems.
\end{itemize}

\begin{figure*}
\centering
    \includegraphics[width=0.97\linewidth]{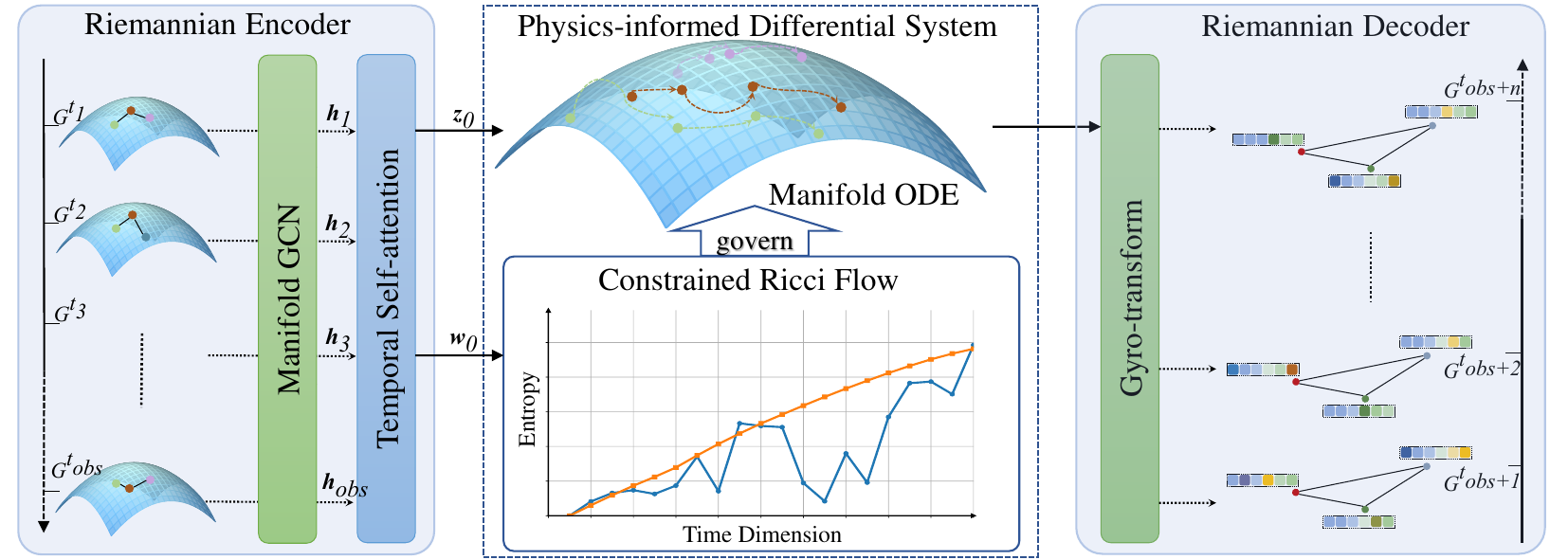}
     \caption{Overall architecture of \textbf{\texttt{Pioneer}}.  }
    \label{Fig-overall}
\end{figure*}

\section{Preliminaries and Notations}

We formally review basic notions of Riemannian geometry and thermodynamics, 
specify the limitations of prior solutions, 
and thereby put forward a more realistic yet challenging problem of \emph{physics-informed dynamic system modeling}.

\paragraph{Riemannian Manifold}
Riemannian geometry provides a systematic approach to study the complex structures. 
In Riemannian geometry, a graph is typically related to some Riemannian manifold, i.e., a smooth manifold $\mathcal M$ endowed with Riemannian metric $\mathfrak g$.
Each point $\boldsymbol x$ in the manifold is associated with a tangent space $\mathcal T_{\boldsymbol x}\mathcal M$, where  Riemannian metric is defined.
Mapping between the tangent space and manifold is given by 
the exponential map $\operatorname{Exp}_{\boldsymbol x}: \mathcal T_{\boldsymbol x}\mathcal M \to \mathcal M$ and logarithmic map $\operatorname{Log}_{\boldsymbol x}:  \mathcal M \to \mathcal T_{\boldsymbol x}\mathcal M$.
There exist three types of isotropic manifolds characterized by the constant curvature $\kappa$: 
hyperbolic space of $\kappa<0$, 
hyperspherical space of $\kappa>0$, 
and Euclidean space of $\kappa=0$ as a special case.

\paragraph{Thermodynamics and Ricci Flow}
In physics, a thermodynamic system is described as a differential heat equation, 
which is equivalently given by Ricci flow from geometric perspective \cite{khan09introRicc}.
\citet{Bai20} define the canonical \emph{Ricci flow} on the graph, 
where the system dynamics is manifested in the evolvement of correlations among nodes, 
i.e., the weight function over time on the edges $w_{ij}(t)$. 
Given a weighted graph $G(\mathcal V, \mathcal E, w(t))$, the Ricci flow is written as 
\begin{equation}
    \frac{d}{dt}w_{ij}(t)=-R(i,j)w_{ij}(t),
    \label{RicciFlow}
\end{equation}
where $\mathcal V$ and $\mathcal E$ denote the node set and edge set, respectively.
\emph{Ricci curvature} $R(i,j)$ in Riemannian geometry characterizes the local geometry on the edge, 
and is defined by the \emph{Wasserstein metric} $W^1$ of measure space regarding the manifold.
There exist two classical discretizations: \citet{Ollivier2007} and \citet{Forman03}.
Ollivier’s  method explicitly calculates the $W^1$ metric where the nested linear programming prevents gradient backpropagation, and thus is not compatible with our deep learning task.
Instead, we employ Forman's version which studies $W^1$ metric implicitly.
Concretely, a family of Forman-Ricci curvature $R^F_{ij}$ is given with some weighting function on the node.
In this paper, we utilize the unity weight for simplicity, and the curvature of the edge $(i,j)\in\mathcal E$ takes the form as follows,
\begin{equation}
  R^F_{ij}=2-\left(\sum_{u\sim i}\sqrt{\frac{w_{ij}}{w_{iu}}}+\sum_{v\sim j}\sqrt{\frac{w_{ij}}{w_{jv}}}\right),
  \label{FormanCurvature}
\end{equation}
where $u$ enumerates the set of vertices that are connected to vertex $i$ (excluding vertex $j$), and same as $v$.

\paragraph{Dynamic System and Graph ODEs} 
We consider a dynamic system of $N$ interacting objects.
The observations consist of objects' trajectories $\mathcal X=\{ \boldsymbol X^1, \cdots, \boldsymbol X^{T_{obs}}\}$ 
and the corresponding interaction graph $\mathcal G=\{ \boldsymbol G^1, \cdots, \boldsymbol G^{T_{obs}}\}$ which evolves continuously.
The system snapshot at timestamp $t$ is given as $(\boldsymbol A^t, \boldsymbol X^t)$, 
where $\boldsymbol A^t(i,j)=w_{ij}^t$ is the weight of the edge between objects $i$ and $j$, and $\boldsymbol X^t\in \mathbb R^{N \times F}$ is the feature matrix of $N$ objects.
In practice, the time intervals are typically nonuniform/irregular.
ODEs show superiority to  model the continuous dynamics, compared to the recurrent models with regular inputs \cite{nips18NODE}.
Concretely, the continuous co-evolvement over  time is described as the ODE with dependent variables of $N$ objects.
The velocity of object is $\frac{d \boldsymbol z_i^t}{d t}=f(\boldsymbol z_1^t, \cdots, \boldsymbol z_N^t)$, 
where $\boldsymbol z$ denotes the coordinate.
Given the initial state $\boldsymbol  z^0$, the entire trajectory is given by
$\boldsymbol  z^T=\boldsymbol  z^0+\int_{t=0}^T f(\boldsymbol z_1^t, \cdots, \boldsymbol z_N^t)$.
Various frameworks with graph ODE have been introduced \cite{kdd21CoupledGraphODE,www24CausalGraphODE,icml23HOPE}.
Surprisingly, we find that \emph{existing solutions expose to the danger of generating geometry-unaware or anti-physics trajectories}, 
e.g., generating trajectories extend beyond the earth surface.
Such limitations motivate us to rethink dynamics modeling of real-world systems and pose the problem 
of physics-informed dynamic system modeling.
In this paper, we focus on the entropy-increasing dynamic system, which is ubiquitous in the real world.
\newtheorem*{def1}{Problem (Physics-informed Dynamic System Modeling)}
\begin{def1}
Given system observations  $\{\mathcal X, \mathcal G\}$, 
our goal is to learn a neural simulator on the underlying geometry $\mathcal M$ of the system, 
which not only captures the system dynamics by predicting future trajectories $X^{T_{obs}+1:T}$, 
but also adheres to the physics law of entropy-increasing.
\end{def1}

\section{Methodology: \texttt{Pioneer}}

We propose a novel physics-informed Riemannian graph ODE for the ubiquitous entropy-increasing dynamic system (\textbf{\texttt{Pioneer}}), 
and our novelty lies in that \emph{the generated prediction of \texttt{Pioneer} is aligned with the  geometry and physics law of real system}.
In \texttt{Pioneer}, we present an ODE system on the Riemannian manifold,
where a \emph{Constrained Ricci Flow} is formulated to control the continuous changes of system states in line with entropy-increasing principle,
and we introduce a \emph{Gyro-transform} for manifold-preserving encoding/decoding, so that the generated trajectories lie on the manifold.
The overall architecture is illustrated in Fig. \ref{Fig-overall}. 
Before elaborating on each component, 
we introduce the Riemannian manifold in which we build our model.

\paragraph{$\boldsymbol \kappa-$sterographical model}
The proposed \texttt{Pioneer} is applicable to any geodesically complete manifold (e.g., hyperbolic space and hyperspherical space) where the closed-form logarithmic map is defined.
We opt for the $\kappa-$sterographical model as it provides the gyrovector formalism to unify constant curvatures of positive, negative and zero values.
Specifically, for any constant curvature $\kappa$ and a dimension $d \geq 2$, 
\emph{$\kappa$-stereographic model} is defined on the smooth manifold of 
$
\mathfrak{S}_{\kappa }^{d}=\{ \boldsymbol{x} \in \mathbb{R}^d | -\kappa \rVert \boldsymbol{x}\lVert^2 < 1 \} 
$
endowed with the Riemannian metric 
$\mathfrak g_{\boldsymbol{x}}^{\kappa } =(\lambda_{\boldsymbol{x}}^\kappa)^2\boldsymbol {I}$, and 
$ \lambda_{\boldsymbol{x}}^\kappa=2(1+\kappa \|  {\boldsymbol{x}} \|^2)^{-1}$ is the conformal factor.
The distance function is thus given as 
\begin{align}
    d_\kappa(\boldsymbol{x}, \boldsymbol{y})=2\tan_\kappa^{-1}(\lVert (-\boldsymbol{x}) \oplus_\kappa \boldsymbol{y} \rVert).
    \label{Distance}
\end{align}
The logarithmic map takes the form of  
\begin{align}
    {\operatorname{Log}_{\boldsymbol{x}}^\kappa}(\boldsymbol{y})
    =\frac{2}{ \lambda_{\boldsymbol{x}}^\kappa }
    \tan_\kappa^{-1}(\lVert (-\boldsymbol{x}) \oplus_\kappa \boldsymbol{y} \rVert)\frac{(-\boldsymbol{x}) \oplus_\kappa \boldsymbol{y}}{\lVert (-\boldsymbol{x}) \oplus_\kappa \boldsymbol{y} \rVert},
    \label{LogMap}
\end{align}
where $\oplus_\kappa$ denotes M\"{o}bius addition of gyrovectors, and we have $\tan_\kappa(x)$ defined as $\tan(x)$ for $\kappa>0$, 
$\tanh(x)$ for $\kappa<0$, and $\tan_\kappa(x)=x$ for  $\kappa=0$.

\subsection{Physics-informed Differential System} 

In the heart of \texttt{Pioneer}, we define a novel differential system on the $\kappa-$sterographical model $\mathfrak{S}^d_\kappa$ to model the underlying dynamics.
The system presents as a couple of ODEs in which a
\emph{Manifold ODE} is governed by the proposed ODE of \emph{Constrained Ricci Flow}, as shown in Fig. \ref{Fig-overall}.
Given the initial state $\boldsymbol{Z}$ on the manifold, the system takes the form of
\begin{align}
\frac{d\boldsymbol{Z}^t}{dt}  &= u(\boldsymbol{Z}^t; \Theta_1)  \in \mathcal T\mathfrak{S}^d_\kappa, \quad  \boldsymbol{Z}^0=\boldsymbol{Z} \label{ManifoldODE}\\
\frac{d{w}_{ij}^t}{dt} & = \left(R^F_{ij}-e^{f(\boldsymbol{z}_i^t, \boldsymbol{z}_j^t; \Theta_2)}\right){w}_{ij}^t, \ \ \ \boldsymbol{z}_i^t, \boldsymbol{z}_j^t \in \mathfrak{S}^d_\kappa. 
\label{NewFlow}
\end{align}
To be specific, 
equation \ref{ManifoldODE} is the \emph{Manifold ODE} on $\mathfrak{S}^d_\kappa$ describing the system co-evolvement over time,
where $\boldsymbol{Z}^t$ collects the objects' latent representations at timestamp $t$.
In the manifold ODE,  
\emph{vector field} $u$ is defined on a section of tangent bundle $\mathcal T\mathfrak{S}^d_\kappa$, 
and assigns latent representation $\boldsymbol z_i^t\in\mathfrak{S}^d_\kappa$ on the manifold to a tangent vector of the bundle.
That is, $u$ is a function that receives manifold-value vector and outputs tangent vector.
In light of the co-evolvement among the objects,
we define the vector field with a graph neural network,
\begin{equation}
u(\boldsymbol{Z}^t; \Theta)=\operatorname{GAT}(\boldsymbol A^t, \operatorname{Log}_{\boldsymbol{o}}^\kappa(\boldsymbol Z^t);\Theta).
\label{VectorField}
\end{equation}
Concretely, we conduct the logarithmic map
$\operatorname{Log}_{\boldsymbol{o}}^\kappa$ to transform manifold-value vector to tangent vector, 
and model the dependent objects in the tangent space with graph attention network ($\operatorname{GAT}$) \cite{nips18gat}, 
where the attentional weight is defined by Eq. 6.
Note that, we cannot directly apply an existing GNN owing to the input/output requirement of the vector field. 

Equation \ref{NewFlow} is the proposed \textbf{Constrained Ricci Flow}, 
an ODE over the manifold-valued $\boldsymbol{z}^t \in \mathfrak{S}^d_\kappa$.
It injects the physics law into the neural system,
while describing the continuous changes of intercorrelation among objects (weighted graph structure).
In particular, 
we propose to constrain the canonical Ricci flow in Eq. \ref{RicciFlow} by a parameterized model,
\begin{equation}
f(\boldsymbol{z}_i^t, \boldsymbol{z}_j^t; \Theta_2)=\sigma(\operatorname{MLP}_{\Theta_2}(\operatorname{Log}_{\boldsymbol{o}}^\kappa(\boldsymbol z_i^t)||\operatorname{Log}_{\boldsymbol{o}}^\kappa(\boldsymbol z_j^t))),
\label{ConstraintFunc}
\end{equation}
where $\sigma$ and  $||$ are sigmoid  and concatenation, respectively. $\operatorname{MLP}$ is a multilayer perception with  parameter $\Theta_2$.
The intuition is that object correlation is expressed by their latent representations,
and the advantage lies in the theory below.

\paragraph{Theory} 
We elaborate on an interesting theoretical result of the Constrained Ricci Flow defined in Eq. \ref{NewFlow}, and we begin with reviewing the definition of von Neumann entropy.
\newtheorem*{def2}{von Neumann entropy \shortcite{MinelloRT19}} 
\begin{def2}
Given a dynamic graph $G$ with evolving weighed adjacency  matrix $\boldsymbol A^t$, 
the von Neumann entropy at time $t$ is defined as $ H^t=-\sum_{i=1}^{n} \lambda_i^t \ln\lambda_i^t$, 
where $\lambda^t_i$ is the $i$th eigenvalue of $\frac{1}{N}\boldsymbol{L}^t$, where 
 $\boldsymbol{L}^t= \boldsymbol{I} -{\boldsymbol D}^{-\frac{1}{2}}\boldsymbol A^t {\boldsymbol D}^{-\frac{1}{2}}$, 
and $\boldsymbol D$ is the diagonal degree matrix.
\end{def2}
\noindent Next, we study the derivative of $\frac{dH}{dt}$ to characterize the continuous change of entropy over time.
\newtheorem*{thm1}{Theorem 3.1 (Entropy Non-decreasing of Constrained Ricci Flow)} 
\begin{thm1}
If the interacting system is governed by the constrained Ricci flow defined in Eq. \ref{NewFlow},
we have the derivative of von Neumann entropy over time $\frac{dH}{dt}\ge 0$  hold under a slight condition that $m\ge n$, where $n$ and $m$ denote the number of nodes and edges, respectively. 
\end{thm1}
\begin{mymath}
\begin{proof}
We present the key ideas and equations here. 
Let $Ric$ to be the summation of Ricci curvature over graph.
First, we begin with analyzing $dRic$ over time under the proposed flow in Eq. \ref{NewFlow}, and give the lower bound as follows.
\begin{align}
    \frac{dRic}{dt}
    &=\frac{1}{2}\sum\nolimits_{e_{ij}}g(e_{ij})\left(g(e_{ij})+e^{f(\boldsymbol{z}_i^t, \boldsymbol{z}_j^t; \Theta_2)}-2\right)\\
    &\geq \frac{1}{2}[\sum\nolimits_{e_{ij}}g(e_{ij})-m]
\end{align}
where $g(e_{ij})=\sum_{u\sim i}\sqrt{\frac{w_{ij}}{w_{iu}}}+\sum_{v\sim j}\sqrt{\frac{w_{ij}}{w_{jv}}}$. 
Second, we scale $\sum_{e_{ij}}g(e_{ij})$ so as to establish the connection between the inequality above and the invariant values of the graph.
\begin{align}
    \sum\limits_{e_{ij}}g(e_{ij})=\frac{1}{2}\sum_{e_{ij}}\sum_{e_{uv}\sim e_{ij}}\left(\sqrt{\frac{w_{ij}}{w_{uv}}}+\sqrt{\frac{w_{uv}}{w_{ij}}}\right) 
\end{align}
\begin{align}
    \sum\nolimits_{e_{ij}}g(e_{ij}) \geq \sum\nolimits_{v}d^2(v)-2m
\end{align}
where $e_{uv}$ enumerate the edges that intersect with edge $e_{ij}$, and $d(v)$ denotes the degree of $v$.
With $\sum_{v}d^2(v)\geq 4m-n$ (detailed in Appendix A.1), we have  $\frac{dRic}{dt}\ge\frac{m-n}{2}$.
Third, we consider the measure space over the Riemannian structure.
Under the second-order derivative w.r.t. time,
the Hessian regarding the entropy is shown to be \emph{k-concavity}, 
yielding the result that changes in entropy and curvature are positively correlated \cite{math09RiccOT},
$ dH \cdot dRic \geq 0$.
Thus, $\frac{ dH}{dt}\ge 0$ holds on the condition of $m\ge n$, completing the proof. 
(Note that, the condition  is satisfied on most of the real world graphs.)
\end{proof}
\end{mymath}
\noindent In other words, dynamic interacting systems endowed with the proposed flow
\emph{are theoretically in line with the generic physics principle of entropy increasing} (more strictly, non-decreasing).
Additionally, it is noteworthy to mention that the canonical Ricci flow do not possess monotonicity over entropy \cite{baptista2024}.





\subsection{Manifold-Preserving Transformation} 
\emph{Aware of the geometry, we consider the manifold-preserving transformation to generate trajectory on the manifold} as well as process system observations.
In previous works, feature transformation typically relies on the couple of exponential/logarithmic maps \cite{nips22QGCN},
introducing a tangent space as the intermediary.
Note that, Euclidean or tangent space is not isometric to the manifold \cite{Petersen16,Yu2022RandomLF}.
It leads to  inevitable mapping error, which is also evidenced in Experiment, and is  a key reason of the fully Riemannian design of \texttt{Pioneer}.


To bridge this gap, we formulate a novel \textbf{Gyro-transform} in any constant curvature.
The advantage lies in that Gyro-transform only requires simple matrix-vector multiplication and does not involves the tangent space (manifold-preserving).
In particular, 
the gyro-transform from $\mathfrak{S}_\kappa^{d_1}$ to $\mathfrak{S}_\kappa^{d_2}$  
is performed by
left multiplying $\boldsymbol z \in \mathfrak{S}_\kappa^{d_1}$ 
with the matrix $Gyr_{\boldsymbol z}(\mathbf W) \in \mathbb{R}^{{d_1}\times{d_2}}$, 
which is defined by rescaling a weight matrix  $\mathbf W \in \mathbb{R}^{{d_1}\times{d_2}}$ as follows,
\begin{equation}
\resizebox{0.47\hsize}{!}{$
    Gyr_{\boldsymbol z}(\mathbf W) = f_{scal}^\kappa(\mathbf W, \boldsymbol z) \mathbf W
$}
\label{GyroTransf}
\end{equation}
where the scaling function $f_{scal}^\kappa$ is defined as
\begin{equation}
\resizebox{0.666\hsize}{!}{$
f_{scal}^\kappa(\mathbf W, \boldsymbol z)=
\frac{\sqrt{\kappa^{-1}[1-(\lambda_z^\kappa-1)^2]}}{\lambda_z^\kappa \lVert \mathbf W\boldsymbol z \rVert},
$}
\label{ScaleFunc}
\end{equation}
$\lambda_z^\kappa$ is the conformal factor, and $sgn$ is the sign function.
The derivation of Gyro-transform is provided in Appendix A.2.
\newtheorem*{thm2}{Theorem 3.2 (Manifold-preserving of Gyro-transform)} 
\begin{thm2}
Given $\boldsymbol z \in \mathfrak{S}_{\kappa }^{d_1}$ and $\kappa$ of either positive or negative value,  
$Gyr_{\boldsymbol z}(\boldsymbol W)\boldsymbol z \in \mathfrak{S}_{\kappa }^{d_2}$ holds for any $\boldsymbol W \in \mathbb R^{{d_1}\times{d_2}}$.
\end{thm2}
\begin{mymath}
\begin{proof}
We verify the correctness of the claim above, and further details are given in Appendix A. 2. 
That is, we are to prove $-\kappa\lVert Gyr_{\boldsymbol z}(\boldsymbol W)\boldsymbol z \rVert <1$ given the domain of manifold $\mathfrak{S}_{\kappa }^{d_2}$.
Instead of the inequality of $\mathfrak{S}_{\kappa }^{d}$, 
we examine the equality in Lorentz/spherical model,
$\mathfrak{L}_{\kappa }^{d}=\{ 
\left[\begin{array}{c}
x_t  \\
\boldsymbol x_s 
\end{array}\right], 
\boldsymbol x_s \in \mathbb R^d
| sgn(\kappa)x^2_t + \boldsymbol x_s^\top\boldsymbol x_s = \frac{1}{\kappa}\}$,
where $sgn$ is the sign function.
(Note that, 
$\mathfrak{L}_{\kappa }^{d}$ model
is equivalent to $\mathfrak{S}_{\kappa }^{d}$ model, 
and the connection is established by $\kappa$-stereographical projection $\Phi: \mathfrak{L}_{\kappa }^{d} \to \mathfrak{st}_{\kappa }^{d}$.)
First, we conduct $\Phi^{-1}$ on $Gyr_{\boldsymbol z}(\boldsymbol W)\boldsymbol z$,
\begin{align}
\resizebox{0.9\hsize}{!}{$
\Phi^{-1}\left(Gyr_{\boldsymbol z}(\boldsymbol W)\boldsymbol z\right)
=\left[\begin{array}{c}
x_t  \\
\mathbf x_s \nonumber
\end{array}\right]
=\left[\begin{array}{c}
\frac{1}{\sqrt{ |\kappa|}}(\lambda_z^\kappa-1)  \\
\lambda_z^\kappa f_{scal}^\kappa(\lambda_z^\kappa)\mathbf W\mathbf z
\end{array}\right]. 
$}
\end{align}
Second, we have the manifold equation of $\mathfrak{L}_{\kappa }^{d_2}$ model 
hold for $\Phi^{-1}\left(Gyr_{\boldsymbol z}(\boldsymbol W)\boldsymbol z\right)$, completing the proof.
\end{proof}
\end{mymath}

\noindent In addition, the proposed Gyro-transform is more computationally efficient than the couple of exponential/logarithmic maps, since $Gyr_{\boldsymbol z}(\boldsymbol W)\boldsymbol z$ has fewer norms and no $\tanh$'s.

\paragraph{Remark} 
Fully manifold operators of \citet{acl22chen,BdeirSL24} are designed for hyperbolic space only,
while Lorentz projector  \cite{aaai23sunli} is  for manifold shift (i.e, from hyperbolic to hyperspherical space, or vice versa).
None of them meets the need of feature transform in any constant curvature.

\subsection{Riemannian Encoder \& Decoder} 

In \texttt{Pioneer}, to generate the initial states of ODEs, the Riemannian encoder is constructed with Gyro-transform and the weighted gyro-midpoint aggregator \cite{Petersen16},
\begin{align}
\resizebox{0.88\hsize}{!}{$
\operatorname{mid}_\kappa({\boldsymbol h},\mathbf{\alpha})
    =\frac{1}{2}\otimes_\kappa
    \left( \sum_{i=1}^{T_{Obs}}
    \frac{\mathbf{\alpha}_i\lambda_{{\boldsymbol h}^i}^\kappa}
    {\sum_{j=1}^{T_{Obs}}\alpha_j(\lambda_{\boldsymbol{h}^j}^\kappa-1)}
    {\boldsymbol h}^i
    \right),
$}
    \label{GyroMidpoint}
\end{align}
where $\alpha_i$ denotes the weight, ${\boldsymbol h}^i \in \mathfrak{S}_\kappa^{d}$, $\operatorname{mid}_\kappa({\boldsymbol h},\mathbf{\alpha}) \in \mathfrak{S}_\kappa^{d}$.
First, for each observed graph $\boldsymbol G^1, \cdots, \boldsymbol G^{T_{obs}}$,  $\boldsymbol G^t(\boldsymbol A, \boldsymbol X)$,
we obtain the object representations by the manifold GCN, where the  feature transformation of Gyro-transform $\boldsymbol h=Gyr_{\boldsymbol x}(\boldsymbol W)\boldsymbol x$ 
is followed up with the aggregation of $\boldsymbol h=\operatorname{mid}_\kappa({\boldsymbol h}; {\boldsymbol a})$.
Second, we conduct the temporal self-attention on the manifold to derive the initial state of manifold ODE in Eq. \ref{ManifoldODE}.
Concretely, the contribution of each timestamp  is modeled with temporal self-attention $\alpha$ parameterized by $\boldsymbol w$,
\begin{align}
\resizebox{0.6\hsize}{!}{$
    \alpha_{i}=\frac{\exp(\boldsymbol w^\top[\boldsymbol t_i||\boldsymbol t_{inital}])}{\sum_j^{T_{Obs}}\exp(\boldsymbol w^\top[\boldsymbol t_i||\boldsymbol t_{inital}])},
$}
\label{AttentionFunc}
\end{align}
\begin{align}
\resizebox{0.886\hsize}{!}{$
        [\boldsymbol t]_{2i}=\sin\left(\frac{t}{10000^{2i/d}}\right),  \ [\boldsymbol t]_{2i+1}=\cos\left(\frac{t}{10000^{2i/d}}\right).
\label{TimeEncoding}
$}
\end{align}
For timestamp $t$, we leverage the time encoding  $\boldsymbol t$ , and  elements of even and odd index are given in Eq. \ref{TimeEncoding}, and $t_{initial}$ is the time of initial state.
Accordingly, the initial state of constrained Ricci flow in Eq. \ref{NewFlow} is given by the distance metric on the manifold,
$
w_{ij}=\frac{
\exp(-d_\kappa({\boldsymbol z}_i, {\boldsymbol z}_j))
}{
\sum_{k=1}^n  \exp(-d_\kappa({\boldsymbol z}_i, {\boldsymbol z}_k))
}.
$

The decoder of \texttt{Pioneer} is given as a Gyro-transform to generate prediction on the manifold.

\paragraph{Objective Function \& Complexity Analysis.}  
Finally, we formulate the reconstruction objective as follows
\begin{align}
\resizebox{0.888\hsize}{!}{$
\mathcal J=\sum\limits_{t=1}^T\left(\sum\limits_{i=1}^N d^2_\kappa\left(\boldsymbol y_i(t), \hat{\boldsymbol y}_i(t)\right) + \sum\limits_{(i,j)\in\mathcal E} \lvert w_{ij}(t)-\hat{w}_{ij}(t) \rvert^2\right),
$}
\label{ObjectiveFunc}
\end{align}
where $\hat{\boldsymbol y}_i(t)$ and $\hat{w}_{ij}(t)$ are given by the observations, and $T$ is the prediction length.
$\boldsymbol y_i(t)$ is the output of the decoder, and $w_{ij}(t)$ is given by the constrained Ricci flow.
We summarize the overall learning procedure  in Algorithm 1.
Accordingly, 
the computational complexity of the encoder is $O(TN^2)$, and $O(TN^2)$ for the proposed ODEs, and $O(N)$ for the decoder,
where $T$ is the number of objects.
It yields the \emph{Square complexity} as a whole.
Though the complexity is the same as existing graph ODEs \cite{kdd23GGODE,aaai24ChenWLLsigned},
\emph{\texttt{Pioneer} is aware of the geometry, 
and the generated system state adhere to the physics law of the ubiquitous entropy-increasing real systems.}

\begin{algorithm}[t]
        \caption{Learning Algorithm of  \texttt{Pioneer}} 
        \KwIn{The system observations $(\mathcal{G}, \mathcal X)$}
        \KwOut{The parameters of \texttt{Pioneer}}
        Initialize modal parameters;

\While{model not converged}{   
            \For{each training sequence} {
                Separate the sequence into observed half $[T_0,T_1]$ and predicted half $[T_1,T_2]$;

\hfill $\rhd$ \emph{Initial State Encoder}\\
                
                 Obtain object representations by manifold GCN;
                 
                 Derive initial states with the temporal self-attention in  Eqs. \ref{GyroMidpoint}-\ref{AttentionFunc};

\hfill $\rhd$ \emph{Riemannian Differential System}\\
                
                Solve the ODEs in Eqs. \ref{ManifoldODE}-\ref{NewFlow} over $[T_1,T_2]$;
               
\hfill $\rhd$ \emph{Decoder}\\
                
                Generate prediction via the Gyro-transform in Eq. \ref{GyroTransf};

            }
            Compute the objective function in Eq. \ref{ObjectiveFunc};

            Update the parameters by gradient descent.
}
\end{algorithm}

\begin{table*}[t]
\centering
\resizebox{1\linewidth}{!}{
\begin{tabular}{c|cc|cc|cc|cc|cc|cc}
\hline
\multirow{3}{*}{\textbf{Methods}}& \multicolumn{6}{c|}{\textbf{Social Dataset (Different Prediction Length)}}& \multicolumn{6}{c}{\textbf{Weather Dataset (Different Prediction Length)}} \\
\cline{2-13}
& \multicolumn{2}{c|}{\textbf{10}}& \multicolumn{2}{c|}{\textbf{20}}& \multicolumn{2}{c|}{\textbf{40}}& \multicolumn{2}{c|}{\textbf{5}}& \multicolumn{2}{c|}{\textbf{10}}& \multicolumn{2}{c}{\textbf{15}} \\
\cline{2-13}
& \textbf{MAPE}& \textbf{RMSE}& \textbf{MAPE}& \textbf{RMSE}& \textbf{MAPE}& \textbf{RMSE}& \textbf{MAPE}& \textbf{RMSE}& \textbf{MAPE}& \textbf{RMSE}& \textbf{MAPE}& \textbf{RMSE} \\
\hline
LSTM    
& 12.18\scriptsize{$\pm$0.21} & 0.14\scriptsize{$\pm$0.04} & 26.16\scriptsize{$\pm$0.25} & 0.26\scriptsize{$\pm$0.05} & 61.74\scriptsize{$\pm$0.35} & 
0.51\scriptsize{$\pm$0.02} & 18.99\scriptsize{$\pm$0.45} & 5.90\scriptsize{$\pm$0.03} & 23.42\scriptsize{$\pm$0.18} & 8.31\scriptsize{$\pm$0.04} & 75.22\scriptsize{$\pm$0.31} & 9.58\scriptsize{$\pm$0.03}  \\
NRI     
& 29.35\scriptsize{$\pm$0.42} & 0.56\scriptsize{$\pm$0.09} & 43.24\scriptsize{$\pm$1.01} & 0.69\scriptsize{$\pm$0.12} & 68.23\scriptsize{$\pm$1.34} & 0.70\scriptsize{$\pm$0.11} & 22.75\scriptsize{$\pm$4.94} & 7.13\scriptsize{$\pm$1.36} & 22.38\scriptsize{$\pm$1.66} & 8.06\scriptsize{$\pm$0.85} & 68.16\scriptsize{$\pm$4.00} & 9.82\scriptsize{$\pm$0.19} \\
VGRNN   
& 11.78\scriptsize{$\pm$0.29} & 0.14\scriptsize{$\pm$0.02} & 26.87\scriptsize{$\pm$0.20} & 0.28\scriptsize{$\pm$0.07} & 54.90\scriptsize{$\pm$0.42} & 0.51\scriptsize{$\pm$0.06} & 14.59\scriptsize{$\pm$0.23} & 5.79\scriptsize{$\pm$0.13} & 19.56\scriptsize{$\pm$1.00} & 7.96\scriptsize{$\pm$0.22} & 74.37\scriptsize{$\pm$3.61} & 9.61\scriptsize{$\pm$0.45} \\
DGCRN   
& 11.87\scriptsize{$\pm$0.20} & 0.14\scriptsize{$\pm$0.02} & 25.84\scriptsize{$\pm$0.23} & 0.31\scriptsize{$\pm$0.06} & 57.03\scriptsize{$\pm$0.35} & 0.58\scriptsize{$\pm$0.09} & 15.39\scriptsize{$\pm$1.36} & \underline{5.75}\scriptsize{$\pm$0.36} & 21.57\scriptsize{$\pm$2.82} & 7.84\scriptsize{$\pm$0.53} & 45.66\scriptsize{$\pm$0.93} & 9.20\scriptsize{$\pm$0.20} \\
NODE    
& 11.64\scriptsize{$\pm$0.17} & 0.16\scriptsize{$\pm$0.02} & 24.92\scriptsize{$\pm$0.35} & 0.30\scriptsize{$\pm$0.07} & 57.50\scriptsize{$\pm$0.33} & 0.64\scriptsize{$\pm$0.03} & 15.96\scriptsize{$\pm$0.16} & 5.99\scriptsize{$\pm$0.15} & 24.37\scriptsize{$\pm$1.80} & 8.22\scriptsize{$\pm$0.76} & 60.92\scriptsize{$\pm$1.35} & 8.83\scriptsize{$\pm$0.13} \\
HBNODE  
& 10.78\scriptsize{$\pm$0.25} & 0.17\scriptsize{$\pm$0.02} & 26.37\scriptsize{$\pm$0.48} & 0.32\scriptsize{$\pm$0.08} & 61.51\scriptsize{$\pm$0.65} & 0.62\scriptsize{$\pm$0.07} & 18.56\scriptsize{$\pm$0.45} & 6.06\scriptsize{$\pm$0.25} & 24.43\scriptsize{$\pm$0.87} & 8.12\scriptsize{$\pm$0.09} & 53.31\scriptsize{$\pm$0.26} & 9.09\scriptsize{$\pm$0.63} \\
LG-ODE  
& 12.80\scriptsize{$\pm$0.16} & 0.13\scriptsize{$\pm$0.01} & 25.46\scriptsize{$\pm$0.25} & 0.24\scriptsize{$\pm$0.04} & 55.59\scriptsize{$\pm$0.43} & 0.51\scriptsize{$\pm$0.03} & 14.53\scriptsize{$\pm$0.26} & 5.78\scriptsize{$\pm$0.11} & 21.99\scriptsize{$\pm$0.66} & 7.86\scriptsize{$\pm$0.42} & 45.41\scriptsize{$\pm$1.35} & \underline{8.68}\scriptsize{$\pm$0.42} \\
CG-ODE  
& 12.63\scriptsize{$\pm$0.20} & 0.13\scriptsize{$\pm$0.02} & 24.47\scriptsize{$\pm$0.38} & 0.23\scriptsize{$\pm$0.04} & 49.09\scriptsize{$\pm$0.35} & \underline{0.49}\scriptsize{$\pm$0.06} & 14.01\scriptsize{$\pm$0.32} & 5.92\scriptsize{$\pm$0.07} & 20.87\scriptsize{$\pm$0.51} & 7.99\scriptsize{$\pm$0.31} & 28.17\scriptsize{$\pm$1.25} & 8.85\scriptsize{$\pm$0.31} \\
HOPE    
& \underline{10.71}\scriptsize{$\pm$0.12} & \underline{0.13}\scriptsize{$\pm$0.01} & \underline{22.13}\scriptsize{$\pm$0.19} & \underline{0.23}\scriptsize{$\pm$0.03} & \underline{47.82}\scriptsize{$\pm$0.22} & 0.49\scriptsize{$\pm$0.03} & \underline{13.76}\scriptsize{$\pm$0.16} & 5.83\scriptsize{$\pm$0.15} & \underline{19.54}\scriptsize{$\pm$0.35} & \textbf{7.75}\scriptsize{$\pm$0.22} & \underline{24.16}\scriptsize{$\pm$0.55} & 8.80\scriptsize{$\pm$0.08} \\
\hline
 \textbf{\texttt{Pioneer}}
& \textbf{9.15}\scriptsize{$\pm$0.12} & \textbf{0.11}\scriptsize{$\pm$0.01} & \textbf{18.67}\scriptsize{$\pm$0.12} & \textbf{0.21}\scriptsize{$\pm$0.01} & \textbf{44.07}\scriptsize{$\pm$0.18} & \textbf{0.45}\scriptsize{$\pm$0.02} & \textbf{12.50}\scriptsize{$\pm$0.48} & \textbf{5.74}\scriptsize{$\pm$0.09} & \textbf{17.27}\scriptsize{$\pm$0.23} & \underline{7.76}\scriptsize{$\pm$0.02} & \textbf{19.39}\scriptsize{$\pm$0.01} & \textbf{8.64}\scriptsize{$\pm$0.03} \\
\hline
\end{tabular}}
\caption{Prediction results with varying prediction lengths on Social and Weather datasets in terms of MAPE(\%) and RMSE. Standard derivations are given in the subscript. The best results are \textbf{bolded} and the runners-up are \underline{underlined}.}
\label{Tab-SocialWeather}
\end{table*}

\begin{table}[t]
\resizebox{1\linewidth}{!}{
\begin{tabular}{p{1.5cm}<{\centering}|p{1.0cm}<{\centering}|p{1.0cm}<{\centering}|p{1.0cm}<{\centering}|p{1.0cm}<{\centering}|p{1.0cm}<{\centering}|p{1.0cm}<{\centering}}
\hline
\multirow{2}{*}{\textbf{Methods}} & \multicolumn{6}{c}{\textbf{CReSIS Dataset (Different Prediction Length)}} \\
\cline{2-7}
& \multicolumn{2}{c|}{\textbf{1}}& \multicolumn{2}{c|}{\textbf{2}}& \multicolumn{2}{c}{\textbf{3}} \\
\hline
& \textbf{MAPE}& \textbf{RMSE}& \textbf{MAPE}& \textbf{RMSE}& \textbf{MAPE}& \textbf{RMSE} \\
\hline
LSTM    
& 3.70 & 52.49 & 4.65 & 66.02 & 4.64 & 68.98 \\
NRI     
& 3.82 & 51.89 & 4.58 & 68.78 & 4.83 & 62.79 \\
VGRNN   
& \underline{3.62} & 50.41 & 4.42 & 65.22 & 4.42 & 62.19 \\
DGCRN   
& 3.71 & 48.52 & 4.41 & 65.02 & 4.38 & 58.18 \\
NODE    
& 3.89 & 52.90 & 4.46 & 65.51 & 4.70 & 63.74 \\
HBNODE  
& 3.76 & 51.78 & 4.41 & 66.24 & 4.73 & 62.75 \\
LG-ODE  
& 3.96 & 52.51 & 4.77 & 63.26 & 4.64 & 61.66 \\
CG-ODE  
& 3.86 & 49.71 & \underline{4.40} & 62.52 & 4.32 & 58.07 \\
HOPE    
& 3.72  & \underline{47.95} & 4.50 & \underline{62.11} & \underline{4.23} & \underline{57.29} \\
\hline
 \textbf{\texttt{Pioneer}}    
& \textbf{3.61}  & \textbf{47.78} & \textbf{4.37} & \textbf{60.86} & \textbf{4.22} & \textbf{57.24} \\
\hline
\end{tabular}}
\caption{Prediction results  with varying prediction lengths on CReSIS datasets in terms of MAPE(\%) and RMSE. The best results are \textbf{bolded} and the runners-up are \underline{underlined}.}
\label{Tab-FloodCresis}
\end{table}

\section{Experiment}




\paragraph{Datasets \& Baselines}
The experiments are conducted on  three real-world  datasets of dynamic systems, 
including Weather \citep{aistats22NikitinWeather}, CReSIS \citep{MMTIMRF13CgogineniCReSIS} and Social \citep{kdd17SGsocial}.
We follow \citet{kdd21CoupledGraphODE} to preprocess the datasets.
To evaluate \texttt{Poineer}, we compare with nine strong baselines of two categories:
(1) RNN-based methods: LSTM, NRI \citep{icml18Kipf}, VGRNN \citep{nips19HVGRNN} and DGCRN \citep{tkdd23DGCRN}.
(2) ODE-based methods: NODE \citep{nips18NODE}, LG-ODE \citep{nips20Huang}, CG-ODE \citep{kdd21CoupledGraphODE}, HBNODE \citep{nips21HBNODE} and HOPE \citep{icml23HOPE}.
So far, existing methods neither consider Riemannian geometry nor system entropy, and we are dedicated to bridging this gap. 

\paragraph{Evaluation Protocol}
We focus on predicting the future trajectories, and evaluate all methods by the metrics of Mean Absolute Percentage Error (MAPE) and Root Mean Square Error (RMSE) following \citep{kdd21CoupledGraphODE,icml23HOPE}.
We perform 10 independent runs for each case, and report the mean with standard derivations. 

\paragraph{Configuration \& Reproducibility}
In \texttt{Pioneer}, graph attention layer in manifold ODE is stacked twice, and $\operatorname{MLP}$ has $1$ hidden layers.
The dimension of latent representation on manifold $\mathfrak{S}$ is set as $16$.
For Euclidean features, we apply the exponential map to obtain corresponding manifold features in the initialization.
The differential system is solved by Euler method on the manifold \cite{riemanEular02}, and parameters are optimized by Adam  with learning rate of $5e-4$.

\begin{table*}[ht]
\centering
\resizebox{0.86\linewidth}{!}{
\begin{tabular}{l|cc|cc|cc|cc|cc|cc}
\hline
\multirow{3}{*}{\textbf{Variants}} & \multicolumn{6}{c|}{\textbf{Social}} & \multicolumn{6}{c}{\textbf{Weather}} \\
\cline{2-13}
& \multicolumn{2}{c|}{\textbf{10}} & \multicolumn{2}{c|}{\textbf{20}} & \multicolumn{2}{c|}{\textbf{40}} & \multicolumn{2}{c|}{\textbf{5}} & \multicolumn{2}{c|}{\textbf{10}} & \multicolumn{2}{c}{\textbf{15}} \\
\cline{2-13}
& \textbf{MAPE} & \textbf{RMSE} & \textbf{MAPE} & \textbf{RMSE} & \textbf{MAPE} & \textbf{RMSE} & \textbf{MAPE} & \textbf{RMSE} & \textbf{MAPE} & \textbf{RMSE} & \textbf{MAPE} & \textbf{RMSE} \\
\hline
$w/oEvo$  
& 11.61 & 0.13 & 24.37 & 0.25 & 59.23 & 0.51 & 15.94 & 6.13 & 18.51 & 7.98 & 20.70 & 8.87 \\
$w/oRic$  
& \underline{10.49} & 0.13 & 22.17 & 0.23 & 53.65 & \underline{0.46} & \underline{12.59} & \underline{5.80} & 17.74 & 7.78 & 19.71 & 8.77 \\
$w/oCon$  
& 10.76 & \underline{0.12} & \underline{18.82} & \underline{0.22} & 54.47 & 0.49 & 12.62 & 5.82 & \underline{17.72} & 7.78 & \underline{19.53} & \underline{8.73} \\
$w/oGyr$  
& 10.63 & \underline{0.12} & 21.58 & 0.23 & \underline{45.09} & \underline{0.46} & 13.63 & 5.93 & 17.74 & \underline{7.77} & 19.75 & 8.84 \\
\hline
 \textbf{\texttt{Pioneer}}
          & \textbf{9.15} & \textbf{0.11} & \textbf{18.67}& \textbf{0.21} & \textbf{44.07}& \textbf{0.45} & \textbf{12.50} & \textbf{5.74}& \textbf{17.27}& \textbf{7.76} & \textbf{19.39}& \textbf{8.64} \\
\hline
\end{tabular}}
\caption{Ablation study results on Social and Weather datasets with varying prediction lengths in terms of MAPE(\%) and RMSE. The best results are \textbf{bolded} and the runners-up are \underline{underlined}.}
\label{Tab-ablation}
\end{table*}

\subsection{Results and Discussion}

\paragraph{Main Results}
Table \ref{Tab-SocialWeather} and Table \ref{Tab-FloodCresis} collect the prediction results on Social, Weather and CReSIS datasets with varying prediction length.
Take CReSIS for instance, the prediction length of $3$ means we examine the prediction error over the future $3$ years.
As shown in the tables, ODEs tend to obtain better performance especially for length term prediction.
Graph ODEs (i.e., HOPE, CG-ODE and LG-ODE) outperform NODE and HBNODE for the most cases, suggesting the importance of correlation modeling.
Note that, the proposed \texttt{Pioneer} consistently achieves the best results on all the real systems except only one case in Weather dateset.
We argue that our success lies in its geometry and the adherence to physics law, and 
we further investigate  proposed components in \texttt{Pioneer}.

\begin{figure} 
\centering 
\subfigure[Social]{
\includegraphics[width=0.482\linewidth]{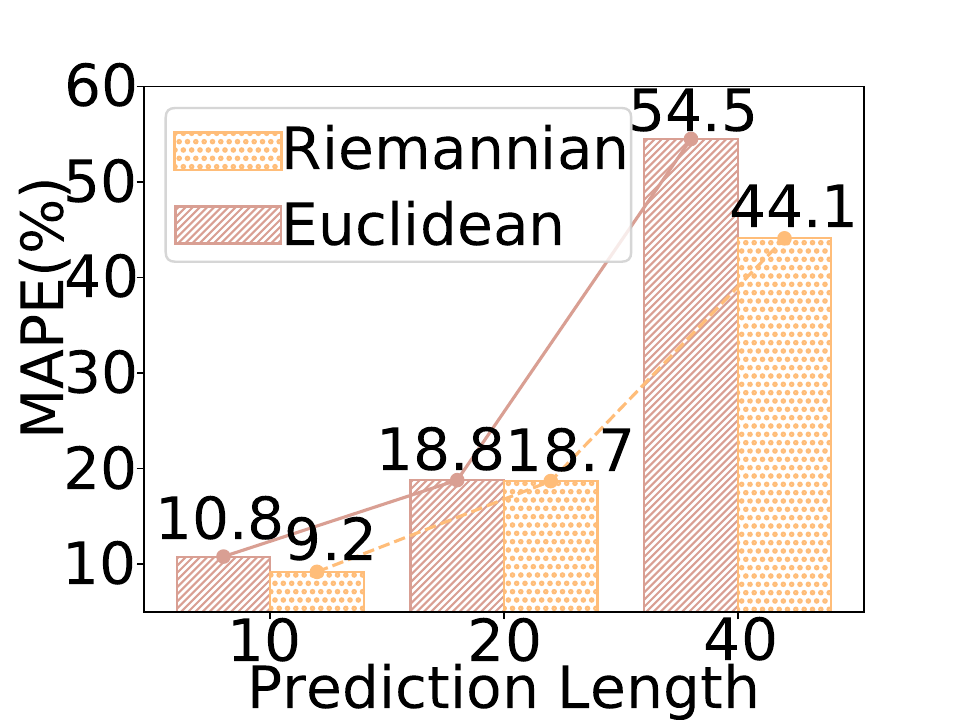}}
\subfigure[Weather]{
\includegraphics[width=0.482\linewidth]{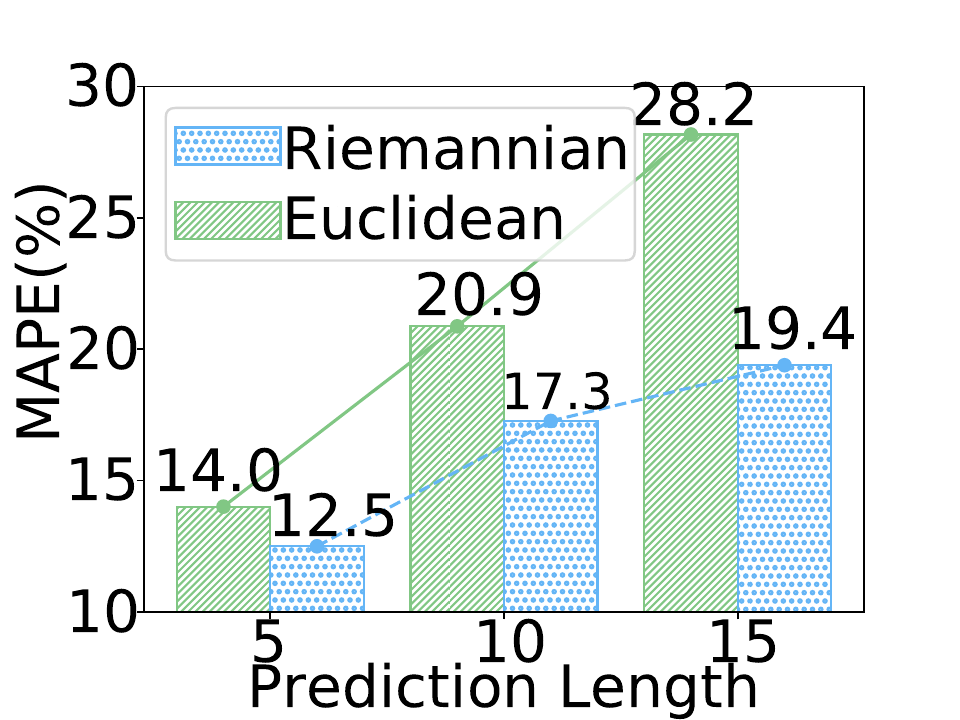}}
\vspace{-0.1in}
\caption{Results of geometric ablation.}
\vspace{-0.1in}
\label{Fig-geometry}
\end{figure}

\paragraph{Ablation Study}
We evaluate the effectiveness of the proposed 1) constrained Ricci flow, and 2) manifold-preserving Gyro-transform.
For the former objective, we design three manifold-valued variants as follows:
$w/oEvo$ variant considers the fixed graph structure and disables Eq. \ref{NewFlow}, verifying the significance of  modeling structural evolvement.
$w/oRic$ variant utilizes the parameterized $f$  over node features to describe the graph structure over time $\frac{dw}{dt}=f$, and does not consider the principle of Ricci flow.
 In $w/oCon$ variant, Eq. \ref{NewFlow} is replaced by the canonical Ricci flow in Eq. \ref{RicciFlow} with neither entropy constraint nor node features.
To evaluate Gyro-transform,  
we design the $w/oGyr$ variant which replaces Gyro-transform by the couple of  exponential and logarithmic maps, involving the tangent space.
The prediction results of the variants are collected in Table \ref{Tab-ablation}, and we find that:
1) \texttt{Pioneer} receives performance gain compared to $w/oGyr$ variant, showing the superiority of manifold-preserving operations.
2) It suggests the importance of considering structural evolvement in dynamic system, given that $w/oEvo$ variant has the worst results.
3) We achieve better results than $w/oRic$ of direct parameterized model and $w/oCon$ of the  canonical  flow, 
while the next part elaborates on another strength of  \texttt{Pioneer}.

\begin{figure}
\centering
    \includegraphics[width=1\linewidth]{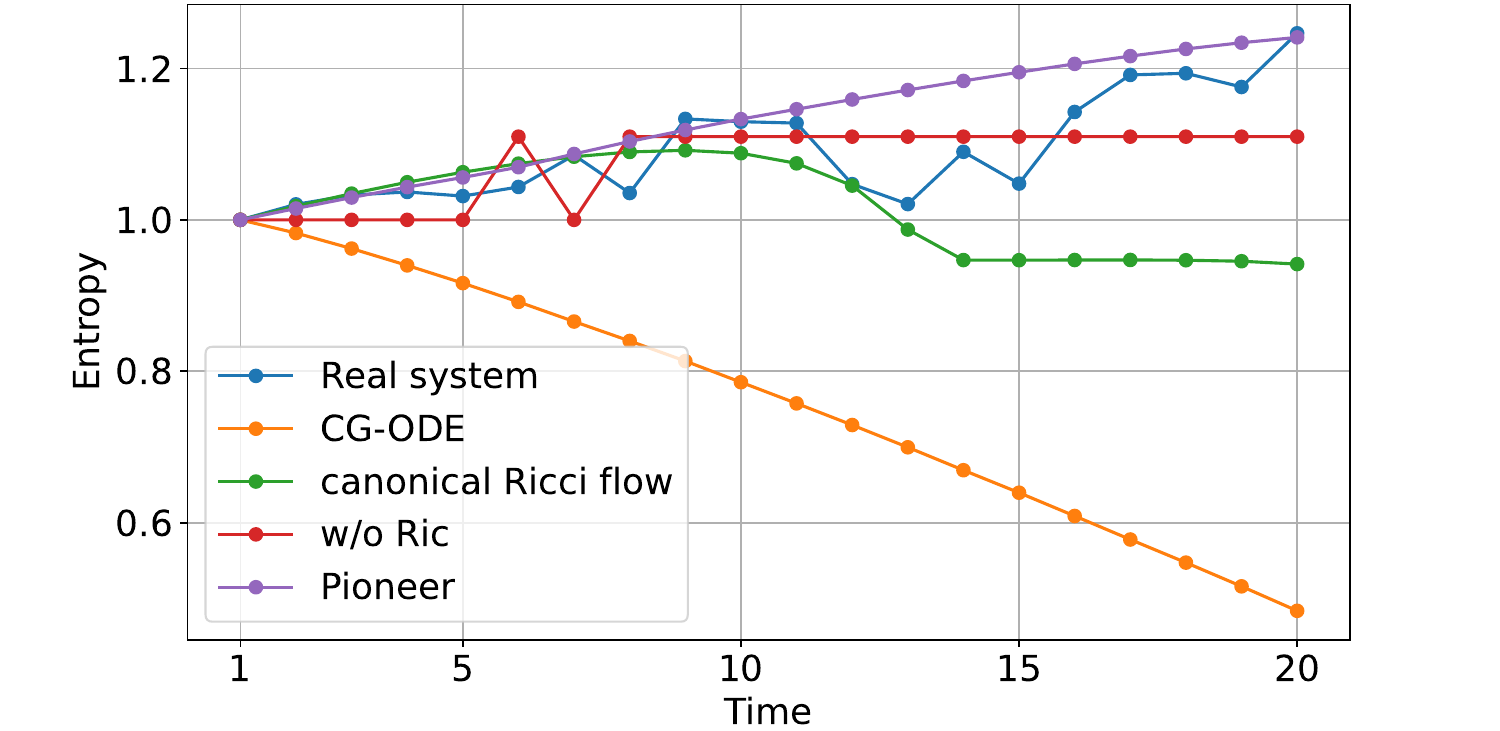}
    \vspace{-0.1in}
     \caption{The changes of entropy over time.}
    \label{Fig-entropy}
\end{figure}

\paragraph{Discussion on Entropy}
We discuss the \textbf{generated system states}  regarding the physics law in dynamic systems.
To this end, 
we examine the continuous entropy changes of the real system and its simulators over time.
The simulators include \texttt{Pioneer}, canonical Ricci flow, CG-ODE \citep{kdd21CoupledGraphODE} as well as $w/oRic$ variant.
Concretely, 
we calculate the von Neumann entropy at different system snapshots, and report the results on Social dataset in Fig. \ref{Fig-entropy}.
The real system presents the tend of entropy increasing in general, 
while the previous CG-ODE violates the fact in physics.
Note that, Ricci flow itself cannot ensure entropy increasing,
and \emph{only the physics-informed
\texttt{Pioneer}  is in line with the  physics fact of entropy increasing principle.}

\paragraph{Discussion on Geometry \& Case Study}



\begin{figure} 
\centering 
\subfigure[\texttt{Pioneer}]{
\includegraphics[width=0.48\linewidth]{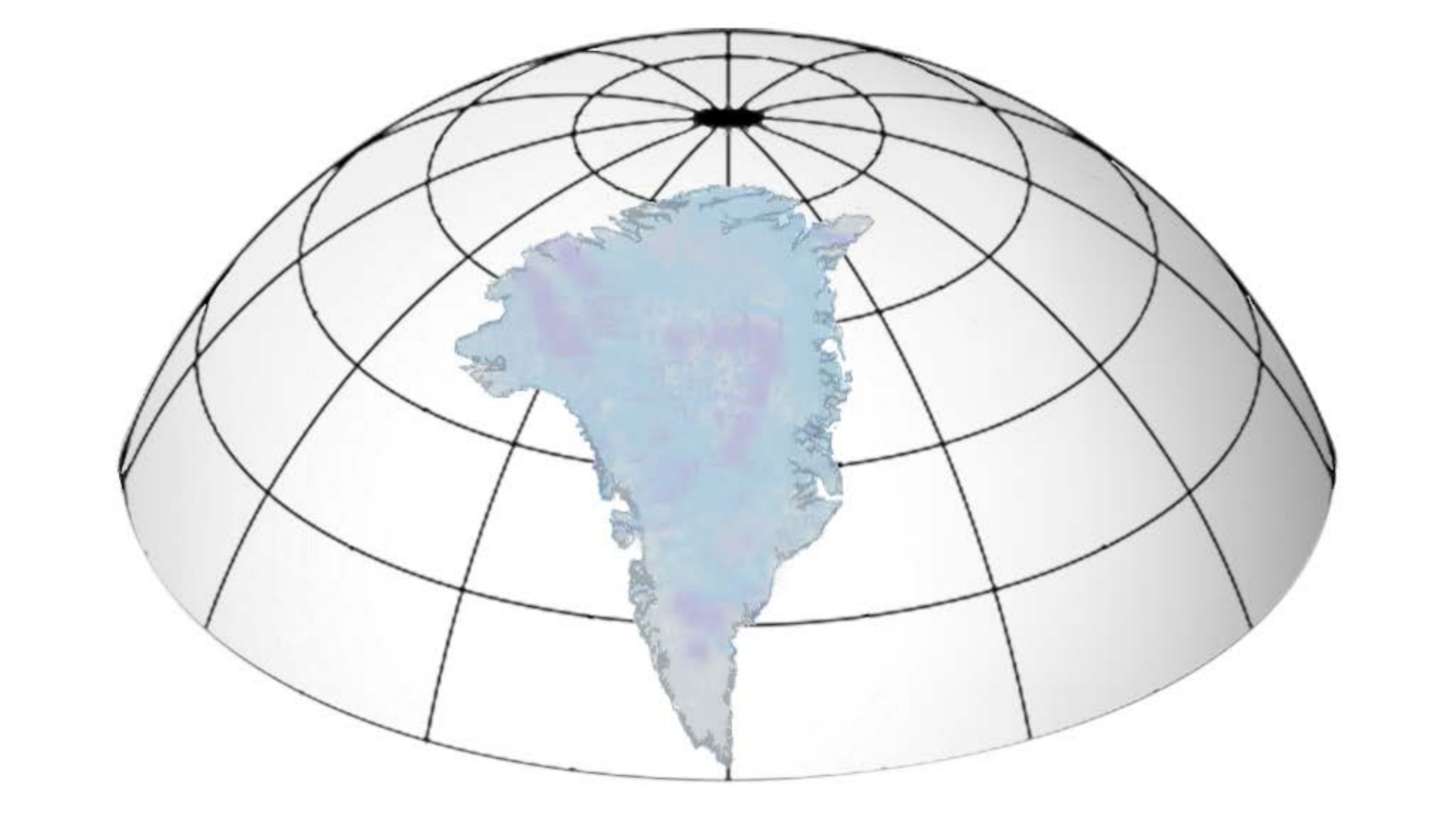}}
\subfigure[Euclidean Modeling]{
\includegraphics[width=0.48\linewidth]{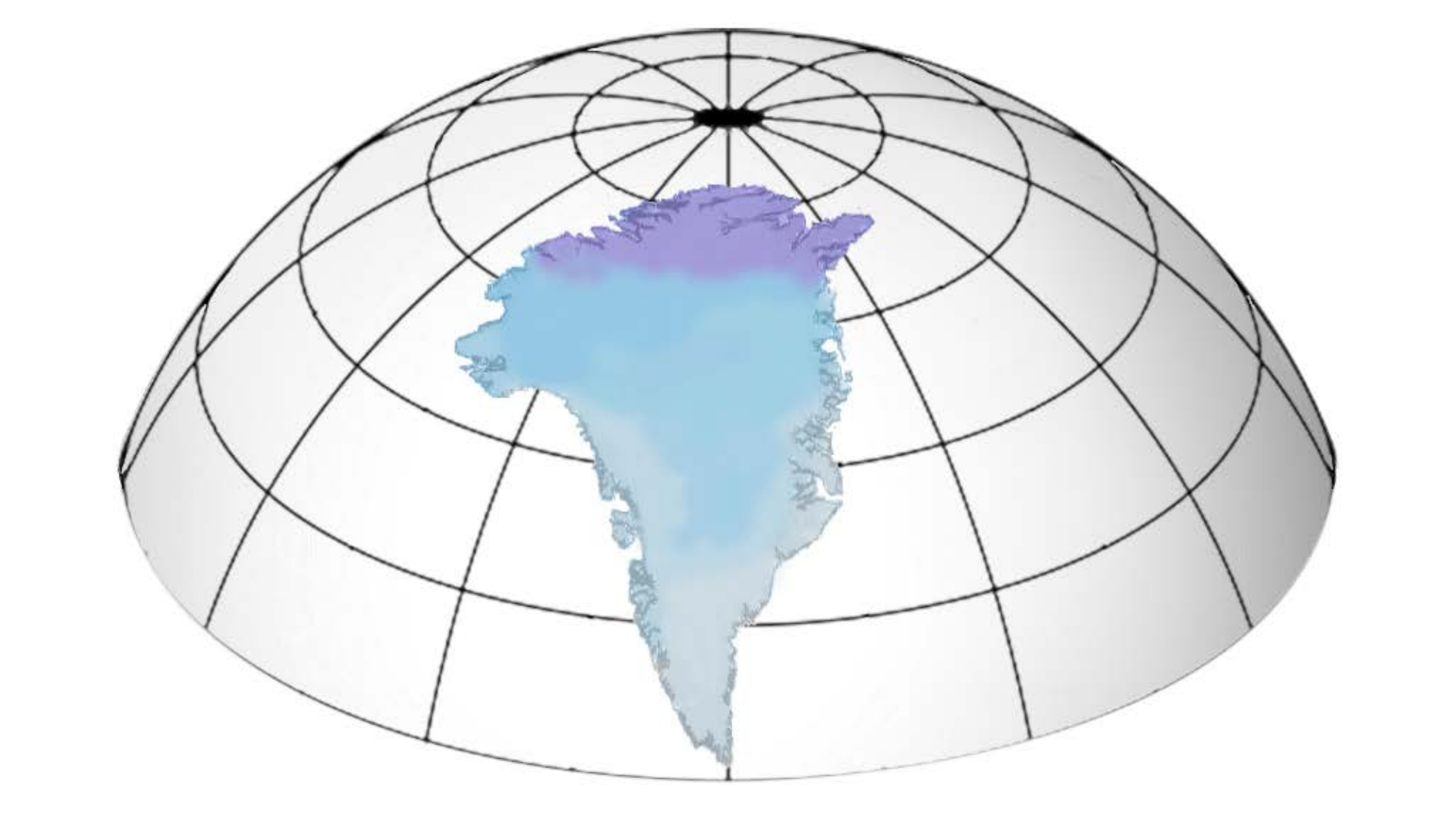}}
\vspace{-0.1in}
 \caption{Prediction results of Glacier thickness.}
\label{Fig-glacier}
\end{figure}

First, we analyze the impact of different geometries with the mean error of prediction results, MAPE metric for instance.
 Fig. \ref{Fig-geometry} shows the comparison between \texttt{Pioneer} and its corresponding Euclidean version.
 We find that the proposed model is superior to the Euclidean one.
Second, we take a closer look by studying the spatial error distribution over the real system.
In particular, we conduct a case study on CReSIS \citep{MMTIMRF13CgogineniCReSIS}, a collection of glacier thickness in Greenland. 
We plot the error distribution of the Riemannian \texttt{Pioneer} on the earth's surface around North Pole in Fig. \ref{Fig-glacier}(a), and that of Euclidean counterpart in Fig. \ref{Fig-glacier}(b), where darker color denotes larger error.
Note that, 
\emph{Euclidean model presents significant error near the North Pole, 
and the error distribution is generally  related to geometric locations},
while \texttt{Pioneer} is more reliable throughout the earth's system, verifying the importance of geometry and the motivation of our study.

\section{Related Work}

\paragraph{Dynamic Interacting System \& Graph ODEs}
Early practices leverage the recurrent models  to study the sequential pattern of the objects.
Given the correlation among objects, GNNs \cite{nips23ssLRD,www24GAUSS} are then incorporated with the recurrent models \cite{nips19rstgnn,nips19HVGRNN,aaai22Fan}.
In recent years, graph ODEs show encouraging results for modeling the correlation and evolvement collaboratively \cite{kdd21CoupledGraphODE,wsdm23Luo}.
\citet{icml23HOPE,kdd22zhang,PMLR20Xhonneux} introduce second-order ODE on graphs,
 while  \citet{kdd21CoupledGraphODE,wsdm23Luo} consider the structural evolvement to model the dynamic correlation, and
\citet{aaai24ChenWLLsigned} reconsider the graph structure of dynamic systems.
Recently, \citet{nips23tango} study  time-reversal symmetric systems. \citet{nips21Lee} present a new parameterization of brackets for learning irreversible dynamics. \citet{nips23Lee} further develop a  graph attention mechanism  to enable the irreversibility in deep GNNs.
Different from previous works, we build the differential system upon Riemannian manifold, considering the entropy increasing of dynamic systems.

\paragraph{Riemannian Deep Learning on Graphs}
Categorized by the geometry, 
hyperbolic space is suitable to hierarchical structures and show the superiority to the Euclidean counterpart \cite{icml24sun,fu24icml,aaai24YCWei}.
Hyperspherical space achieves remarkable success to embed cyclical structures \cite{icml23sphereFourier}.
Recent studies further investigate the constant curvature spaces \cite{icml20kGCN,nips24sun}, product spaces \cite{iclr19Gu,aaai22sunli,aaai24sunli}, quotient spaces \cite{nips22QGCN}, SPD manifolds \cite{icml23GyroSpace}, etc. 
For dynamic graphs, there exist Riemannian models focusing on link prediction and/or node classification \cite{kdd21yang,aaai21sunli,cikm22sunli,aaai23sunli}.
Recently, Ricci curvature is introduced to structure learning \cite{icml23revisitRicci,icdm23sunli} or clustering \cite{ijcai23sunli}, while its physics aspects are not explored.
Also, we notice that \citet{nips20mainfoldode,www24sunli,sigir24sunli,WangSun24cikm,iclr24FlowMatch} studies Riemannian ODEs or SDEs as generative models.


\section{Conclusion}
We present the first graph differential system on Riemannian manifold (\texttt{Pioneer}) to study the dynamics of real-world systems.
In accordance to the physics law, \texttt{Pioneer} is endowed with the  constrained Ricci flow, which is theoretically proved to obey entropy non-deceasing with slight constraints.
Meanwhile, we formulate the Gyro-transform to guarantee the manifold-preserving in both encoding and decoding processes.
Finally, we empirically discuss the geometry and entropy changes in real system.

\newpage

\section{Acknowledgements}
This work is supported in part by NSFC under grants 62202164 and 62322202.
Prof. Philip S. Yu  is supported in part by NSF under grants III-2106758, and POSE-2346158.

\bibliography{aaai25}

\end{document}